# PtychoFM: A Foundation Model for Ptychographic Imaging

## Authors

Aileen Luo[1,*], Emon Dey[2], Ming Du[1], Tao Zhou[3], Peco Myint[1], Himanshu Goel[4], Shashank Subramanian[5], Xiangyu Yin[1], Tekin Bicer[1,6], Siniša Veseli[1], Cheng-Chu Chung[4,7], Jeffrey A. Klug[1], Aldo Peña Pérez[8], Llorenç Fanals-i-Batllori[8], Christopher Kenney[8], Henry Shi[1], Lorenzo Rota[8], Angelo Dragone[8], Zirui Gao[4], Douglas A. Bennett[9], John A. B. Mates[9], Leila R. Vale[9], Michael R. Vissers[9], Yi Jiang[1], Junjing Deng[1], Tejas Guruswamy[1], Varuni K. Sastry[10], Bowen Zheng[11], Takenori Shimamura[4], Oliver Hoidn[12], Aashwin Mishra[12], Hannah Parraga[1], Brandon Stone[1], Jonathan Baldwin[1], Aaron Michelson[13], Michael Wojcik[1], Xiaoyin Zheng[1,6], Robert Underwood[2], Yijiang Li[2], Jinxing Jiang[1], Nicholas Schwarz[1], Apurva Mehta[12], Johanna Nelson Weker[14], Hassina Bilheux[15], Joel N. Ullom[9], Alexander Hexemer[11], Antonino Miceli[1], Kibaek Kim[2], Hanfei Yan[4], Mathew J. Cherukara[1,*], Alec R. Sandy[1], Stefan Vogt[1], Laurent C. Chapon[1]

## Affiliations

[1]Advanced Photon Source, Argonne National Laboratory, Lemont, IL 60439, USA

[2]Math and Computer Sciences Division, Argonne National Laboratory, Lemont, IL 60439, USA

[3]Center for Nanoscale Materials, Argonne National Laboratory, Lemont, IL 60439, USA

[4]National Synchrotron Light Source II, Brookhaven National Laboratory, Upton, NY 11973, USA

[5]National Energy Research Scientific Computing Center, Lawrence Berkeley National Laboratory, Berkeley, CA 94720, USA

[6]Data Science and Learning Division, Argonne National Laboratory, Lemont, IL 60439, USA

[7]Department of Materials Science and Chemical Engineering, Stony Brook University, Stony Brook, NY 11794, USA

[8]Technology Innovation Directorate, SLAC National Accelerator Laboratory, Menlo Park, CA 94025, USA

[9]Quantum Electromagnetics Division, National Institute of Standards and Technology, Boulder, CO 80305, USA

[10]Argonne Leadership Computing Facility, Argonne National Laboratory, Lemont, IL 60439, USA

[11]Advanced Light Source, Lawrence Berkeley National Laboratory, Berkeley, CA 94720, USA

[12]Linac Coherent Light Source, SLAC National Accelerator Laboratory, Menlo Park, CA 94025, USA

[13]Center for Functional Nanomaterials, Brookhaven National Laboratory, Upton, NY 11973, USA

[14]Stanford Synchrotron Radiation Laboratory, SLAC National Accelerator Laboratory, Menlo Park, CA 94025, USA

[15]Neutron Scattering Division, Oak Ridge National Laboratory, Oak Ridge, TN 37830, USA

*aileenluo@anl.gov, mcherukara@anl.gov

## Abstract

Ptychography provides high-resolution lensless imaging across X-ray, electron, and optical microscopy, but reconstructing images from diffraction measurements generally requires computationally intensity iterative phase retrieval. These methods often require experiment-specific analysis, limiting real-time feedback as data acquisition rates increase. Here, we demonstrate PtychoFM, a physics-aware foundation model for ptychographic imaging that combines large-scale pre-training with experiment-specific fine-tuning to reconstruct complex object images directly from individual diffraction patterns. Trained using a differentiable forward model without object-space labels, PtychoFM transfers from step-scan training data on one sample to fly-scan measurements of other samples with different feature sizes, morphologies, and refraction indices. At a synchrotron beamline, diffraction frames were streamed from the detector to a single GPU for inference at up to 200 frames per second and returned for real-time visualization. PtychoFM provides a practical route to rapid, transferable ptychographic reconstruction and live experimental feedback under nonideal conditions.

## Main Text

### Introduction

Ptychography is a scanning coherent diffractive imaging technique that circumvents the resolution limits of conventional lens-based microscopy by numerically retrieving the phase of the scattered wavefield[1,2]. A coherent beam is scanned across an extended object with overlapping illumination regions; the redundancy in the measured far-field diffraction patterns enables joint recovery of both the complex object and probe. The technique has enabled breakthrough imaging across synchrotron X-ray nanoscience[3,4,5,6], electron microscopy[7,8], and optical coherent imaging[9,10], with applications including imaging of biological specimens at near-atomic resolution[11,12] and three-dimensional imaging of integrated circuits when combined with projection-based techniques such as laminography[13] or tomography[14,15]. Bragg geometry extensions further enable 3D strain imaging in extended crystals[16,17,18].

A central challenge in interpreting ptychographic data is the ill-posed phase retrieval problem: diffraction measurements record only intensities, and the phases of the scattered waves must be inferred from the overlap redundancy in the dataset. The modern framework for iterative phase retrieval traces its origins to the seminal algorithm of Gerchberg and Saxton[19], which alternated enforcement of known amplitude constraints between real and reciprocal space to retrieve the phase of an electron wave. The idea of using lateral diversity in a scanning experiment to overdetermine the phase problem was proposed by Hoppe[1] and later formalized by Bates and Rodenburg[20]. Conventional algorithms – including the extended ptychographic iterative engine[21] (ePIE), regularized ptychographic iterative engine[22] (rPIE), least-squares maximum likelihood[23,24] (LSQML), alternating projection-based difference map[25] (DM), relaxed averaged alternating reflections[26] (RAAR), and proximal operator-based algorithms[27] – address this through iterative updates of the object and probe with the objective of either minimizing the simulation-measurement mismatch or finding the intersection of multiple constraint subspaces. Methods with adaptive step size and preconditioning mechanisms, represented by LSQML, often offer superior robustness to noise and more quantitatively accurate reconstructions. While highly effective, iterative phase retrieval is computationally intensive, often requiring hundreds to thousands of iterations per dataset to converge, and are highly sensitive to hyperparameter tuning and initialization quality. This computational demand is an increasingly critical bottleneck as modern synchrotron and electron microscopy facilities generate data at rates of hundreds of diffraction patterns per second, far outpacing real-time analysis capacity.

Machine learning has achieved significant improvements in ptychographic reconstruction speed, with supervised convolutional neural network (CNN) based approaches[28,29,30] demonstrating fast, high-quality recovery of complex objects across X-ray and electron modalities. Nonetheless, these CNN regressors are limited to narrow experimental conditions and cannot generalize to varying probe types or samples. Recurrent[31] and unrolled-iteration architectures[32,33] improve sample diversity but retain the sequential structure of classical iterative solvers, diminishing their speedup benefit over traditional algorithms. Physically informed CNNs[34,35,36] leverage automatic differentiation to build the ptychographic forward model into the network architecture to improve generalizability, yet none have demonstrated direct transferability across samples with vastly different indices of refraction.

Vision transformers (ViTs)[37] offer a compelling alternative architecture for phase retrieval. The global receptive field provided by multi-head self-attention[38] is naturally suited to the non-local Fourier relationship between real space structure and reciprocal space diffraction. Nevertheless, ViT adaptation faces challenges on small datasets because the inductive biases of convolutions, such as locality and translation equivariance, are beneficial priors that pure attention architectures lack[39]. Strategies such as polar attention[40] and preservation of the relative scan positions during training[41] have been developed specifically for ptychography, but do not demonstrate generalizability across experimental datasets.

Here, we introduce PtychoFM, a physics-aware foundation model framework for ptychographic imaging designed to move beyond reconstruction models trained for a single sample, probe, or experimental configuration. Rather than training a separate reconstruction model for each experimental configuration, PtychoFM uses a pre-training and fine-tuning strategy that allows a single flexible architecture to potentially adapt to diverse X-ray ptychography instruments while retaining the speed of direct neural-network inference. The model incorporates the physics of coherent diffractive imaging into the reconstruction pipeline, providing a physical inductive bias without requiring an iterative solver at inference. We show that this approach enables fast, high-resolution reconstruction of samples unseen during training and demonstrate live deployment at a synchrotron beamline, where PtychoFM is fine-tuned on step-scan data and then used for inference on fly-scan ptychography measurements. We further demonstrate that PtychoFM can recover meaningful reconstructions from a dataset where conventional iterative algorithms fail because the measurements violate a central assumption of standard ptychography: that the probe is invariant across the scanned region. These results position PtychoFM as a scalable, practical route toward real-time, transferable ptychographic reconstruction for next-generation coherent imaging instruments.

## Results

### Physics-aware model design

PtychoFM maps individual coherent diffraction patterns directly to the complex transmission function of the illuminated sample region. As shown in Figure 1a, diffraction measurements are encoded by a vision transformer (ViT) into a latent representation that captures their global structure and is decoded by convolutional neural networks (CNNs) into complementary amplitude and phase images. These outputs define a complex object estimate that is passed through a differentiable ptychographic forward model with the known probe. The re-computed diffraction pattern is then compared with the measurement, and the resulting error is used to update the network weights. Unlike conventional ptychography, which derives redundancy from known probe positions and overlapping scan points, this single-shot coherent diffractive imaging (CDI) formulation uses the finite probe illumination as the support constraint for each measurement. The architecture is therefore governed not only by learned image representations, but also by wave-propagation physics that restricts the space of admissible reconstructions.

The complete workflow (for details, refer to Methods, PtychoFM architecture and training) links large-scale pre-training, experiment-specific adaptation, and real-time deployment (Fig. 1b). Pre-training uses a large, simulated dataset (Methods, Simulated data generation) generated from diverse object structures and probes reconstructed from experimental data from multiple instruments at the Advanced Photon Source, the National Synchrotron Light Source II, and the Linac Coherent Light Source. These probes span distinct focusing optics, including Fresnel zone plates[42], multilayer Laue lenses[43], and Kirkpatrick-Baez mirrors[44]. The pre-trained model is then aligned to a new experiment using a comparatively small step-scan dataset, from which iterative reconstruction (Methods, Iterative phase retrieval) is used to obtain the complex probe function. Once fine-tuned, PtychoFM performs inference (Methods, Edge inference) on diffraction patterns streamed directly from the detector to an edge GPU device with the model weights preloaded onto it[45], providing a path from acquisition to reconstruction and live visualization without running an iterative phase retrieval optimization for every measurement.

**Performance on experimental data**

Having established a workflow that combines broad pre-training and specific fine-tuning (Methods, PtychoFM architecture and training) for adaptation to the target experimental distribution, we tested the generalization of the resulting model beyond the sample class and acquisition mode used for fine-tuning. The base model was fine-tuned on spiral step-scan ptychography data (Methods, Experimental data acquisition) acquired from a 130 nm CMOS technology integrated circuit sample (see Supplementary Information, SI Fig. S1) and was subsequently evaluated on fly-scan[46] data from a 28 nm CMOS technology sample with markedly different feature sizes and device morphology. Diffraction frames were streamed directly from the detector to the PtychoFM model deployed on a single GPU at the beamline for live inference at acquisition rates of up to 200 frames per second, and the predicted object patches were placed at their positions for real-time visualization (Fig. 1b; Methods, Edge inference; Supplementary Video 1). Figure 2 compares the phase predicted by PtychoFM with reconstructions from five iterative algorithms (Methods, Iterative phase retrieval): difference map (DM), relaxed averaged alternating reflections (RAAR), extended ptychographic iterative engine (ePIE), regularized PIE (rPIE), and least-squares maximum likelihood (LSQML). Because an exact object is not available for experimental data, a high signal-to-noise-ratio (SNR) LSQML reconstruction serves as the common “pseudo ground truth” reference (Fig. 2a). This reference provides a stringent benchmark for object contrast, phase morphology, and small feature recovery, while avoiding the circularity of comparing against simulated targets.

The reconstruction methods are benchmarked on low SNR (collected at 200 Hz, 5% of the exposure time of the high SNR reference) fly-scan data, where reduced photon statistics and continuous beam motion across the sample can degrade the diffraction measurements and introduce motion-blurring effects[47]. The comparison was performed on matched fields of view and evaluated using phase contrast, contrast-normalized root-mean-square error (RMSE), and the Pearson correlation coefficient (PCC). We characterize the phase contrast by the range from the 1$^{st}$ to 99$^{th}$ percentiles to capture dynamic distribution while excluding outliers. Contrast-normalized RMSE measures pixelwise disagreement after accounting for method-dependent phase contrast and constant phase offset, whereas PCC measures agreement in spatial variation independently of positive linear scaling. Lower RMSE and higher PCC indicate closer agreement with the high SNR LSQML reconstruction. The corresponding amplitude prediction is provided in the SI (Fig. S2). Across the full field of view (Fig. 2b), PtychoFM outperforms DM, RAAR, and ePIE, while LSQML and rPIE reconstructions agree most closely with the reference.

The observed ranking may be explained in part by how the algorithms respond to noisy, mutually inconsistent measurements. DM and RAAR use projection-and-reflection updates to seek simultaneous agreement with the measured diffraction amplitudes and with a common object and probe. Low photon counts, fly-scan blur, and partial coherence may make these constraint sets inconsistent, so the aggressive updates can amplify structured artifacts or diverge. ePIE is more stable because it transfers each magnitude-substitution residual to the object and probe through relaxed incremental updates. Nevertheless, it still enforces each noisy measured amplitude and normalizes its updates by the maximum probe intensity, which can produce conflicting corrections, slow updates in less illuminated areas, and eventually stagnate. PtychoFM instead supplies an implicit learned image prior constrained by the forward model, which may mitigate the ill-posedness of the problem due to measurement-level inconsistencies and account for its better low SNR RMSE and PCC relative to DM, RAAR, and ePIE. This interpretation is consistent with the strong image-denoising performance demonstrated by deep neural networks[48].

Conversely, rPIE regularizes its incremental updates according to the local probe intensity, retaining a denominator floor in dim regions while permitting larger, better-conditioned updates where the data are informative[22]. LSQML minimizes a dataset-level least-squares likelihood rather than seeking exact satisfaction of every magnitude measurement and, in the implementation used here, additionally employs illumination preconditioning, compact batching, and momentum. These mechanisms provide greater tolerance to inconsistent low SNR measurements and offer potential explanations for the lower full-field RMSE and higher PCC of rPIE and LSQML relative to PtychoFM, even though the region of interest comparison (Fig. 2c, d) reveals more structure artifacts in their reconstructions, emphasizing a limitation of ranking methods from global summary metrics alone. Although the local RMSE and PCC ordering varies among methods, PtychoFM exhibits fewer periodic texture artifacts on the low SNR data than any of the iterative algorithms, including rPIE and LSQML. These artifacts originate from poor recovery of the probe and periodic scan positions[25]. Effects from the grid scan are present in the PtychoFM prediction as well, however, they originate from the stitching of the individual object patches. The cleaner local background and more consistent device morphology with the reference show that PtychoFM preserves localized structure despite the change in data acquisition mode and sample.

Edge-response analysis on a different region of the 28 nm CMOS technology sample with a strong edge feature (Fig. S3) provides an additional comparison of effective spatial resolution. Resolution is calculated as the full width at half maximum (FWHM) of the Gaussian obtained by differentiating an error-function fit to the same phase edge in each reconstruction. The nominal FWHM is 85.5 ± 2.8 nm for PtychoFM, 41.2 ± 8.9 nm for LSQML, and ranges from 31.1 to 56.9 nm across the other iterative methods. The narrower edge widths of the iterative reconstructions should be interpreted cautiously because their edge profiles contain highly non-monotonic slopes and reconstruction artifacts (Fig. S3c). We note that this edge-based analysis is an effective sample- and imaging-system resolution rather than an isolated measure of reconstruction resolution; it also depends on the intrinsic width and geometry of the sample feature, scan position errors, measurement noise, and smoothing introduced during reconstruction. Furthermore, resolution of single-frame CDI solely depends on the momentum transfer vector $\boldsymbol{Q}$-range of the diffraction, whereas ptychography relies also on the spatial overlap information, inherently limiting PtychoFM's prediction resolution compared to conventional approaches. The full suite of metrics is tabulated in the SI (Table S1, S2) and details on their calculation are found in Methods, Metrics.

To evaluate whether the pre-training and fine-tuning workflow transfers to other ptychography instruments, we applied PtychoFM to ptycho-tomography data from an integrated circuit sample measured at the Hard X-ray Nanoprobe[49,50] of the National Synchrotron Light Source II. The pre-trained model was fine-tuned on diffraction data from a -20° projection (Fig. S4) and then applied to a 60° projection (Fig. 3, S5), testing inference on a different projected sample structure. Figure 3 compares the predicted phase with a reference reconstruction obtained using a multi-stage alternating direction method of multipliers (mADMM)[27]. In this case, both mADMM and PtychoFM reconstruct the same measured diffraction data, providing a direct comparison of reconstruction methods as opposed to the separately acquired, higher SNR reference of Figure 2. PtychoFM reproduces the principal projected device features across the full field-of-view (Fig. 3a), with lower contrast-normalized RMSE and higher PCC in the selected region of interest (Fig. 3b). The stronger local agreement indicates that PtychoFM accurately reconstructs high-frequency scattering corresponding to fine spatial features, while the full-field comparison also reflects differences in the broader phase distribution and the phase contrast at the right-hand sample boundary. These results support the applicability of the pre-training and fine-tuning workflow across facilities and demonstrate transfer across projection angles within an experiment, extending its use to real-time visualization of projections acquired for ptychographic tomography.

**Practical advantages over iterative phase retrieval**

We find that PtychoFM generalizes not only across object morphology, but also to samples with substantially different complex refractive index to its fine-tuning data distribution. The model enables direct inference reconstruction of datasets with very little scattering variation, under experimental conditions which are difficult for iterative algorithms. Figure 4 compares PtychoFM inference with different LSQML settings on a Josephson junction[51], a circuit element used to make superconducting quantum interference devices. This sample is primarily niobium-based, with a device thickness of ~1 μm mounted on 100 μm of unpolished silicon. We note that the model checkpoint used in Figure 4 is different from the one used in Figure 2 because the two datasets were collected in different experiments (see Table S3 for summary of experimental conditions) but is similarly fine-tuned only on data from the 130 nm CMOS integrated circuit sample (a primarily copper-based, ~10 μm thick sample on 40 μm of polished silicon).

Standard LSQML settings with compact batching and momentum (Methods, Iterative phase retrieval) do not produce a stable joint object and probe reconstruction for the Josephson junction dataset. Figure 4b shows a visually selected reconstruction checkpoint at 70 epochs, where LSQML recovers recognizable object phase morphology and incorrect amplitude, with substantial grid artifacts in both channels. Continuing the optimization (Fig. 4c, 300 epochs) results in an incorrect object, despite decreasing reconstruction loss (Fig. S6d). The first probe mode (insets, Fig. 4b, c) shows that the reconstructed probe appears to disperse with additional iterations, which we know to be implausible given the successful reconstruction of other samples from the same experiment. Nevertheless, probe retrieval cannot simply be bypassed by fixing a probe from a different sample; object-only optimization of the Josephson junction also fails to recover the object amplitude and phase (Fig. S6b). These incomplete reconstructions are not specific to the LSQML algorithm, as ePIE likewise does not recover a stable, physically plausible object and probe (Fig. S6a).

A successful iterative reconstruction requires a scan-position-dependent flat-field illumination correction of the data to remove the moving background produced as the zone plate was scanned through a non-uniform incident illumination (see Methods, Iterative phase retrieval and Fig. S7 for details on data preprocessing), followed by total variation regularization (Fig. 4d). Evidence that correction of the background illumination is necessary is provided by comparing the recovered probes under standard and advanced LSQML settings, which show that the non-uniform background is being absorbed into the probe estimate, thereby violating the fundamental mathematical assumption in ptychography that the probe must be invariant throughout the extent of the measurement. Here, orthogonal probe relaxation[52], which models variation of the probe by positionally-dependent linear superpositions of a small number of eigenmodes, cannot accommodate the changes in the probe (Fig. S6c). This iterative result demonstrates that the dataset is not categorically unreconstructable, but its dependence on specialized preprocessing and regularization illustrates the value of real-time prediction using PtychoFM, as this solution would be difficult to achieve during an experiment. Furthermore, this is also a transiently recovered state, albeit with improved object amplitude and phase and reduced artifacts, emphasizing that simply monitoring loss is insufficient to gauge reconstruction quality and hyperparameter tuning.

In contrast, PtychoFM recovers the object amplitude and phase by direct inference from routinely cropped, binned, and normalized diffraction data, without bespoke flat-field correction or regularization (Fig. 4a). Iterative reconstruction repeatedly updates the object and probe to reduce disagreement between measured diffraction and the predictions of a physical forward mode. When experimental conditions violate the model assumptions, these updates can absorb unmodeled contributions into the object or probe, as observed here for the moving illumination background. PtychoFM uses the forward model during pre-training and fine-tuning, but inference applies the learned mapping without further optimizing a diffraction-space loss or explicitly enforcing consistency with the measured intensities. The learned prior may therefore allow recognizable object structure to be predicted from imperfect measurements without forcing the prediction to account for contributions that the assumed physical model cannot describe. Although iterative reconstruction can be adapted to nonideal data, as the corrected LSQML result demonstrates, doing so requires first diagnosing the cause of failure and then selecting and tuning an experiment-specific strategy, which is difficult to determine in advance and impractical to optimize during data acquisition. Once fine-tuned for an experiment, PtychoFM applies a common inference workflow across different samples and acquisition conditions without redesigning the reconstruction strategy for each dataset. It therefore can provide immediate evidence that a difficult dataset contains recoverable structure and can guide subsequent investment in higher-fidelity iterative analysis. This capability is particularly valuable at the present beamline, where scanning the zone plate through a non-uniform incident beam causes background features to move relative to the diffraction patterns, violating the position-independent probe assumption of standard ptychography. By producing interpretable reconstructions under this nonideal geometry, PtychoFM provides a practical route to live ptychographic imaging without hardware modification, while remaining complementary to iterative reconstruction.

## Discussion

### Model scaling

The PtychoFM framework establishes a path toward a foundation model for ptychography. Thus, in addition to evaluating the model performance on the downstream task of object reconstruction, we also explore the training behavior through scaling of the model and training data sizes. Following the approach used to derive compute-optimal language model scaling laws by Hoffmann *et al.*[53], we use isoFLOP profiles to determine how a fixed training compute budget should be allocated between model capacity and optimization duration (Fig. 5). Models spanning approximately 3 million to 724 million parameters were evaluated at target budgets from $6 \times 10^{17}$ to $10^{19}$ FLOPS (floating-point operations per second). The model configurations are tabulated in the SI (Table S4). Within each budget, larger models received fewer optimization steps (and consequently, data tokens) so that total compute remained approximately constant. We then fit a quadratic to log validation loss as a function of log model size and used its minimum to estimate the compute-optimal model size. The corresponding number of processed patch tokens was interpolated within the same compute slice.

Across the measured range, the fitted optimal model size and optimal token count both increase approximately as power laws with training FLOPS (Fig. 5b). This trend is evident up to $10^{19}$ FLOPS, but the $10^{19}$ isoFLOP profile does not have a sufficiently well-resolved minimum to support a reliable curve-fit optimum. We therefore regard that fit as inconclusive and do not extrapolate it. The model size profile provides clear evidence that increasing model size to approximately 40 million parameters (the PtychoFM checkpoints reported in this work are about 43 million parameters) improves diffraction-space loss, but no strong evidence that increasing capacity beyond this scale improves performance under the compute conditions tested.

The token profiles also show reduced loss from making multiple passes over the same data. Computer vision models benefit greatly from training over many epochs of repeated data: for example, a ViT trained for 300 epochs outperforms a corresponding 90-epoch baseline on ImageNet-1k[54]. This behavior differs from the usual scaling setting in large language models (LLMs), in which additional training tokens are generally treated as additional data; controlled data-constrained LLM experiments find that repeated tokens have diminishing value as the number of epochs increases[55]. Our results nevertheless agree qualitatively with the central finding of Hoffmann *et al*.[53] that training a smaller model on more tokens can be preferable to undertraining a larger model. For PtychoFM, revisiting a diffraction pattern is not equivalent to receiving a new independent example, but it can continue to optimize consistency with the known forward model.

This distinction also illustrates the value of incorporating physics into the model architecture. The ptychographic forward model constrains the set of admissible amplitude and phase predictions, reducing the amount of unique training data required to train a ViT compared with learning an arbitrary inverse mapping from labeled ground truth image supervision alone. The resulting scaling trends should therefore be treated as empirical guidance for this physics-constrained vision task rather than as directly transferable LLM scaling laws. Moreover, diffraction-space loss is an incomplete measure of model performance and can only screen candidate models; the ultimate criterion is whether a model reconstructs the object faithfully on experimental measurements, judged in object space against appropriate experimental references and physical consistency checks. Based on those metrics, a 200 million parameter model trained with the same order of magnitude of FLOPS as the deployed 43 million parameter model shows some potential benefit to increasing model size even with reduced data (Fig. S8, Table S5); however, the tradeoff is additional computational cost during fine-tuning and inference. Extending these scaling experiments through additional orders of magnitude would be necessary to determine whether a larger compute-optimal model emerges, but such experiments are beyond our present computational resources.

**Limitations**

Although PtychoFM demonstrates comparable reconstruction quality to standard iterative algorithms in a fraction of the time, the model has several limitations. PtychoFM presently requires fine-tuning for each experiment because changes in illumination, photon energy, detector geometry and sampling, and measurement noise span a data space that pre-training cannot fully cover. The fine-tuning data therefore needs to reflect the intended deployment conditions and should be acquired with an equal or shorter exposure time, corresponding to comparable or lower SNR, than the anticipated test data. A model finetuned on 0.1s exposure data reconstructs data acquired at 0.01s exposure; however, a second fine-tuning stage on a smaller 0.01s exposure dataset improves reconstruction at the higher scan rates (Fig. S9, Table S6, Supplementary Video 2), suggesting that fine-tuning at lower SNR or across a range of photon counts may improve robustness.

PtychoFM also cannot recover the probe or refine scan positions because it treats each diffraction pattern as an independent single-shot coherent diffractive imaging measurement, with a known probe during training. Positions are used only to place and average predicted patches, without enforcing agreement in their overlaps. This independent treatment can be advantageous in generalization of sample morphology prediction but also limits phase quantitativeness. The bounded phase ranges used to generate the pre-training data (Methods, Simulated data generation) may bias the learned mapping toward the phase contrast represented in that distribution. More fundamentally, the diffraction-space loss is insensitive to a constant phase offset within each predicted patch, even with a known probe. Any common phase convention is therefore implicit in the learned mapping rather than established by an explicit calibration between patches. Averaging overlapping predictions without aligning their phase offsets or enforcing their consistency may attenuate spatial phase variations and reduce the dynamic range of the stitched object. Thus, agreement in local morphology does not by itself establish quantitative recovery of phase contrast across the full field of view. The absence of joint object-probe optimization and overlap constraints can nevertheless be advantageous when the measurements are inconsistent with the assumed probe or scan geometry, as in the Josephson junction experiment. Hybrid ML and real-time iterative[56] approaches could therefore retain the speed and robustness of single-shot inference, which can be combined with phase correlation-based position refinement[57], while addressing changes in the probe when experiment conditions allow.

**Live reconstruction at scale**

The practical benefits of PtychoFM inference become clearest at large scale. Figure 6a shows a phase prediction spanning approximately 0.5 mm $\times$ 0.2 mm, assembled from 250 scans (over 10 million diffraction patterns) acquired over approximately 27 hours. Iterative reconstruction of a dataset of this size would be burdensome not only because of its aggregate data volume and computational cost, but also because the non-uniform illumination problem identified in Figure 4 becomes more severe across the larger scan area. For each 100 $\mu$m $\times$ 100 $\mu$m region, the optics module is translated over the full range of its piezoelectric motors before the sample is repositioned with coarse motors. The resulting variation in the illumination and background across many optical and sample positions further violates the constant-probe assumption made in conventional ptychographic reconstruction. In this specific experiment, this problem is exacerbated by an earlier beam drop that caused the beam path to shift horizontally (Fig. S10a) and brief, intermittent instabilities in the beam over the course of extended measurement (Fig. 6b). PtychoFM bypasses this potential failure mode and allows an instrument whose operation is not optimal for ptychography to produce useful large field-of-view imaging.

Even at instruments designed specifically for ptychography, reconstructing a large area at the highest possible fidelity is often unnecessary. A rapid, noisy survey is sufficient to identify interfaces, devices, defects, or other regions of interest, after which additional measurements and reconstructions can be targeted. PtychoFM can reconstruct such low SNR survey data as they are collected, giving the scientist an immediate spatial map for deciding where higher quality data are worth the additional acquisition and reconstruction cost. Additionally, this capability can support machine-interpretable experimental context needed for autonomous workflows[58], in which vision language models and agents select subsequent scan regions and coordinate data analysis.

These results establish PtychoFM as a practical framework for ptychographic imaging under low-signal and experimentally nonideal conditions. Unlike iterative methods, which solve a new optimization problem for each dataset and may require dataset-specific choices of preprocessing, physical models, constraints, and stopping criteria, PtychoFM produces real-time reconstructions from individual diffraction patterns after experiment-specific fine-tuning. The model is complementary to, rather than a replacement for, iterative phase retrieval, which remains essential for probe recovery and high-fidelity quantitative reconstruction. Instead, PtychoFM provides immediate and consistent, physics-constrained feedback that can identify recoverable structure, guide subsequent measurements, and prioritize more detailed analysis. This combination of speed, adaptability, and beamline integration provides a foundation for large field-of-view imaging and autonomous ptychography workflows.

## Methods

### PtychoFM architecture and training

The network, implemented in PyTorch, takes a single-channel 256 x 256 diffraction amplitude as input. Before encoding, intensities are transformed as $\log_{10}(I)$, which compresses the dynamic range of the diffraction pattern while retaining low-intensity structure. A convolutional patch embedding layer with kernel size and stride 16 converts the image into 256 non-overlapping patches arranged on a 16 x 16 lattice. Each patch is projected to a 512-dimensional token and added to a learned positional embedding. The token sequence is processed by 12 pre-norm transformer blocks, each comprising 8-head self-attention and a feed-forward multilayer perceptron with expansion ratio 4.0. After layer normalization, the patch tokens are reshaped directly into a 512-channel, 16 x 16 spatial feature map.

Amplitude and phase are reconstructed with separate convolutional decoders that share the same topology but not weights. Each decoder contains four upsampling stages, matching the 16 x 16 transformer output to the 256 x 256 object grid. At each stage, two transposed-convolution blocks with 3 x 3 kernels are followed by bilinear upsampling by a factor of two. The base channel width is 64 and batch normalization and dropout (0.1 probability) are applied within the decoder blocks. Both decoders use a hyperbolic tangent output activation function followed by multiplicative scaling to constrain their respective outputs to physically relevant ranges. The resulting real-valued amplitude and phase images define the complex object $O = Ae^{i\phi}$.

The model is trained to optimize the object in an unsupervised manner; the predicted complex object is inserted into the ptychographic forward model:

$$I = \sum_{m} \|\mathcal{F}(O \cdot P_m)\|^2$$

The object estimate is multiplied by the complex probe $P_m$ (assumed known) to form the exit wave, which is propagated to the far-field detector plane via a Fourier transform $\mathcal{F}$. The magnitude of the propagated wave is squared to obtain intensity. Intensities are summed over the configured incoherent probe modes $m$ and normalized on a per-sample basis before conversion to diffraction amplitude. The training target is the measured diffraction amplitude; thus, gradients are driven by disagreement in the detector plane rather than by an object reconstruction target.

We used the adaptive momentum (ADAM) optimizer[59] with separate parameter groups for the encoder and each decoder. This separation permits component-specific learning rates and provides additional optimization flexibility, particularly during fine-tuning. Pre-training was performed on 50 million simulated diffraction patterns (Methods, Simulated data generation) for 20 epochs using an inverse-weighted mean-squared-error (MSE) loss, which increases the relative contribution of weak high-frequency scattering features. The pre-training learning rate was $10^{-5}$, with a global batch size of 32,768. Pre-training was distributed across 256 nodes with 4 NVIDIA A100 GPUs per node on the Perlmutter supercomputer at the National Energy Research Scientific Computing Center. For experiment-specific adaptation, the model was initialized from the pre-trained weights and fully fine-tuned without parameter freezing on approximately 50,000 diffraction patterns for 101 epochs. During fine-tuning, the loss function was changed from inverse-weighted MSE to an unweighted mean absolute error to improve training stability, because the experimental diffraction patterns did not contain high-$Q$ scattering corresponding to small sample features. Fine-tuning used a learning rate of $10^{-4}$ for a global batch size of 3,072 across three nodes with eight NVIDIA A100 GPUs per node. Live fine-tuning at the beamline used a learning rate of $10^{-4}$ for a global batch size of 3,200 across 50 nodes with 4 NVIDIA A100 GPUs per node on the Polaris machine at the Argonne Leadership Computing Facility through a Globus[60,61] compute workflow.

**Simulated data generation**

Complex object functions were generated from natural images in ImageNet[62]. Each image was cropped to 550 x 500 pixels and used to define the spatial structure of both the object magnitude and phase. For each object, a lower magnitude bound $A_{LB}$ was sampled from a uniform distribution $U(0.8, 0.98)$ and the image values were rescaled to span the interval $[A_{LB}, 1]$. Phase range $\phi_{max} - \phi_{min}$ was independently from $U(0.8, 1.8)$ and centered at zero. A uniform 30 x 30 grid, corresponding to 900 probe positions, was generated for each complex object.

Illumination functions were drawn from a library of experimentally reconstructed probes acquired at multiple synchrotron and X-ray free-electron-laser facilities. Probe selection used weights based on the focusing optic to compensate for the strong imbalance in the library, which contains orders of magnitude more Fresnel zone plate probes than multilayer Laue lens or Kirkpatrick-Baez mirror probes. Consequently, each individual zone-plate probe had a lower probability of selection than a probe from the less represented optic classes. The selected probe was further augmented by applying a random defocus with magnitude up to 100 μm. The complex object, augmented probe, and scan positions were then passed to the Pty-Chi (see Methods, Iterative phase retrieval) forward propagator to simulate the diffraction intensities used for pre-training.

**Iterative phase retrieval**

Reconstructions were performed with the PyTorch-based iterative phase retrieval software Pty-Chi[63,64]. Diffraction frames were loaded from raw scan files and cropped to the selected region of interest (256 x 256 pixels) with the scattering pattern centered and hot pixels masked out. This basic preprocessing ensures consistency with the real-time model inference workflow. The processed diffraction patterns were paired with the corresponding probe positions, which were converted from meters to object pixel coordinates using the object pixel size given by $\lambda d / Np$, where $\lambda$ is the probe wavelength, $d$ is the detector distance, $N$ is the number of detector pixels, and $p$ is the detector pixel size. Step-scan reconstructions use calculated theoretical positions without position correction, whereas fly-scan reconstructions use calculated positions with a uniform random perturbation up to 10% of the step size with position correction to mitigate grid artifacts from perfectly periodic scan positions[25]. At the beginning of each experiment, a single scan was reconstructed for 500 iterations using the GPU implementation of ePIE in PtychoLib[65]. The probe recovered from this preliminary reconstruction was supplied to all Pty-Chi engines as the initial probe guess. The object was initialized as a single slice uniform complex transmission over a support chosen from the measured positions and probe extent, with an additional object margin.

Projection-based reconstructions used the difference map (DM) and relaxed averaged alternating reflections (RAAR) engines. The DM reconstructor follows the ptychographic formulation of Thibault *et al.*[25]; it applies the measured amplitude projection to the reflection of the retained exit wave about the object-consistency projection and relaxes the resulting exit wave update. The RAAR implementation follows the exit wave update and overlap-normalized object and probe projections described by Marchesini *et al*[66]. In this work, both projection engines were restricted to a single probe mode due to GPU memory constraints (an example of a probe reconstructed with multiple modes is provided in the SI Fig. S11, where it is apparent that partial coherence is not required). The probe was rescaled once to match the integrated intensity of the active diffraction dataset, after which its power constraint was disabled. Probe retrieval used an inertia value of 0.5 and object retrieval in RAAR additionally used an inertia of 0.5. Object-probe scale ambiguity was removed every 10 epochs. Probe positions were refined by gradient correction (treated differently from gradient-based reconstruction engines in Pty-Chi) during the first three epochs while the probe remained fixed, after which position refinement stopped and probe refinement began. DM used an exit wave update relaxation of 0.5, whereas RAAR used a reflection weight $\beta = 0.5$.

The extended ptychographic iterative engine (ePIE) algorithm in Pty-Chi follows the original ePIE formulation of Maiden and Rodenburg[21] and the rPIE implementation follows Maiden *et al.*[22]. The current exit waves of each batch are propagated to the detector plane, where their magnitudes are replaced with the measured diffraction amplitudes, and the residuals are propagated back to the sample plane to form the analytical object and probe update directions. In ePIE, these directions are weighted by the conjugate probe or object patch and normalized by the corresponding maximum intensity; object and probe updates use step size 0.1 and $\alpha = 1$. In rPIE, the denominators combine the local intensity with the corresponding maximum intensity, limiting large updates in weakly illuminated regions; object and probe updates use step size 1 and $\alpha$ = 0.1.

LSQML reconstructions were performed with the least-squares maximum-likelihood engine in Pty-Chi following Odstrčil *et al*[24]. The LSQML configuration uses stochastic gradient descent parameter updates with object and probe step sizes set to 1, Gaussian noise modelling, compact batching, and a momentum-acceleration gain of 0.5, with the compact batch and momentum implementation following Wakonig *et al*[67].

The Josephson junction sample could not be reconstructed using the default configurations described above due to its weak scattering under the experimental conditions, as well as the specific instrument design of the beamline. The zone plate, rather than the sample, was scanned through a nonuniform incident illumination. Consequently, the background intensity pattern caused by the upstream beam moved relative to the diffraction pattern across scan positions, violating the position-independent illumination assumed in ptychographic reconstruction. Thus, additional preprocessing of the diffraction patterns to correct for the background illumination was required to reconstruct those scans. Using the calculated scan positions transformed to the zone plate plane, each diffraction amplitude pattern was translated to its corresponding position in the laboratory coordinate system, thereby aligning the background features across the scan. A mean diffraction amplitude was also calculated from the untranslated patterns in the scan coordinate system to average out the moving background and was then translated to each scan position. The flat-field illumination was estimated by dividing the sum of the position-registered measured amplitudes by the sum of the correspondingly translated mean amplitudes. This flat-field illumination was translated back into the coordinate system of each zone-plate position and divided from the associated measured amplitude. Additionally, during the reconstruction, a total variation constraint was applied to the complex object estimate using two iterations of the Chambolle algorithm[68,69] with weight 0.01, beginning at epoch 20, and repeated every 5 epochs. This regularization suppressed pixel-scale noise while preserving sharp object boundaries.

**Edge inference**

Edge deployment followed the real-time streaming ptychography workflow reported by Babu and Zhou *et. al*[45]. After fine-tuning PtychoFM, the selected PyTorch checkpoint was exported to ONNX without the ptychographic forward model and compiled as a half-precision TensorRT engine on an NVIDIA RTX 6000 Pro GPU at the beamline. During acquisition, detector frames (cropped and/or binned) were streamed over the local network as an EPICS (Experimental Physics and Industrial Control System) process variable to an inference server. The inference server normalized each diffraction pattern by an estimated average maximum detector count value, scaled the normalized frame to 10000 counts (to match the intensity level in training), and transformed the intensities as $\log_{10}(I)$. The preprocessed pattern was passed to the preloaded TensorRT engine to predict its amplitude and phase. The inference server then published the predicted object patch and server status on dedicate channels. For live visualization, each predicted object patch was cropped to a square that fit within the probe footprint, indexed onto a full object array according to its theoretical calculated position, and averaged with overlapping predictions to update a stitched image.

**Experimental data acquisition**

Experimental ptychography data were acquired at Sector 26 (Hard X-ray Nanoprobe) of the Advanced Photon Source using X-rays with a photon energy of 9.659 keV. Far-field diffraction patterns were recorded with an Eiger2 1M detector with a pixel size of 75 μm positioned 1 meter downstream of the sample. The illumination was formed by a Fresnel zone plate focusing optic and the sample was placed 100 μm downstream of the focal plane, producing a probe with fitted FWHM of approximately 520 nm. The optics module, and therefore the probe, was raster scanned across the stationary sample.

Fine-tuning data were collected from a 130 nm CMOS technology sample using spiral step scans with an exposure time of 5 ms and mean overlap ratio of 0.87, estimated by the one minus the nearest neighbor distance divided by the probe diameter. Test data were acquired from a different 28 nm CMOS technology sample in fly-scan mode along a zigzag grid trajectory. During this acquisition, the motors moved continuously and the detector was triggered to acquire frames with an exposure time of 5 ms at a nominal step size of 93.59 nm (14 pixels on the reconstructed object) for a nominal overlap ratio of 0.83. The high SNR reference reported in Figure 2 was collected as a grid of spiral step scans with an exposure time of 100 ms, where each individual spiral scan had a mean overlap ratio of 0.87 (overlap between scans was filtered out at the time of reconstruction to maintain relatively even sampling). The experimental details for the Josephson junction sample differ slightly and are reported in the SI (Table S3) for brevity.

**Metrics**

The contrast range was defined as the difference between the 1$^{st}$ and 99$^{th}$ percentile values from each reconstruction (to exclude outliers). Each reconstruction was normalized by its own contrast range. An additive offset was then applied to the normalized reconstruction so that its mean residual relative to the independently normalized reference was zero; this offset is the value that minimizes the subsequent mean squared error. The reported contrast-normalized root mean squared error (RMSE) was the square root of the mean squared residual over all retained pixels. Pearson correlation coefficient (PCC) was calculated between the unscaled reconstruction and reference values over the same retained pixels. Because PCC is invariant to constant offsets and positive linear scaling, it complements the pixelwise RMSE by measuring correlated spatial structure.

Spatial resolution was estimated from the edge response of each phase reconstruction. Eleven evenly spaced parallel line profiles were sampled across the same sharp edge in each image over a 10-pixel-wide strip (6.685 nm reconstructed pixel size), with each profile spanning the 300 nm fitting interval length interpolated to 600 points. For each reconstruction, the pointwise mean of these profiles was fitted with the error function model:

$$f(x) = A\,\mathrm{erf}\left[\frac{(x-x_0)}{\sigma\sqrt{2}}\right] + C;$$

where $A$ describes half the edge contrast, $x_0$ is the edge position, $C$ is the mean of the two plateau levels, and $\sigma$ describes the edge width. The derivative of this error function is a Gaussian with standard deviation $\sigma$; the spatial resolution was therefore calculated as the full width at half maximum of that Gaussian, $2\sigma\sqrt{2\ln 2}$, or approximately $2.355\sigma$. The reported coefficient of determination ($R^2$) compares the residual sum of squares between the sampled profile and the fitted error function with the total sum of squares about the profile mean, quantifying the fraction of profile variance explained by the fit (values closer to one indicate that the error function better represents the local edge profile).

## Data availability

An experimental dataset that supports the findings of this study will be made available in a public repository upon acceptance of this manuscript for publication.

## Code availability

The code and trained model developed in this study will be made available in a public GitHub repository upon acceptance of this manuscript and approval from our institution.

## Acknowledgements

This work is supported in part by the U.S. Department of Energy's Genesis Mission and the Office of Science, Office of Advanced Scientific Computing Research's ModCon under Contract No. DE-AC02-06CH11357. Work performed at the Center for Nanoscale Materials and Advanced Photon Source on APS beam time awards (DOI: https://doi.org/10.46936/APS-192289/60015799, https://doi.org/10.46936/APS-192595/60015990), both U.S. Department of Energy Office of Science User Facilities, was supported by the U.S. DOE, Office of Basic Energy Sciences, under Contract No. DE-AC02-06CH11357. This research used resources of the National Energy Research Scientific Computing Center (NERSC), a Department of Energy User Facility using NERSC award ALCC-ERCAP0038260. This research used resources of the Argonne Leadership Computing Facility, which is a U.S. Department of Energy Office of Science User Facility operated under contract DE-AC02-06CH11357. This research used resources at the 3-ID Hard X-ray Nanoprobe beamline of the National Synchrotron Light Source II, a U.S. Department of Energy (DOE) Office of Science User Facility operated for the DOE Office of Science by Brookhaven National Laboratory under Contract No. DE-SC0012704. This research used resources of the Center for Functional Nanomaterials (CFN), which is a U.S. Department of Energy Office of Science User Facility, at Brookhaven National Laboratory under Contract No. DE-SC0012704.

## Author contributions

A.L. designed, implemented, trained, and tested the model, and collected and analyzed the data, with advising from M.D., T.Z., and M.J.C. E.D., S.S., P.M., V.K.S., Y.L., and R.U. scaled the model training on NERSC Perlmutter and ALCF Aurora. H.G., J.A.K., Y.J., O.H., J.D., and M.D. prepared experimental data from the various facilities. M.D. and P.M. curated experimental probes and simulated the training data. P.M., H.G., B.Z., C.C., O.H., A. Mishra, and X.Z. contributed to model testing. X.Y., S.V., T.Z., and T.G. implemented the edge inference workflow. T.B., E.D., M.D., and H.P. implemented the Globus compute fine-tuning workflow. A.P.P, L.F., C.K., H.S., L.R., A.D., and A. Miceli designed the 130 nm and 28 nm CMOS samples, as well as the integrated circuit sample measured at the NSLS-II. D.A.B., J.A.B.M., L.R.V., M.R.V., and J.N.U. designed and fabricated the Josephson junction sample. T.S., A. Michelson, C.C., and H.Y. acquired and analyzed the data at NSLS-II HXN. C.C., H.G., and T.S. deployed the model for fine-tuning and inference at HXN along with P.M., A.L, and H.Y. Z.G. created the custom preprocessing for iterative reconstruction of the Josephson junction sample. B.S., J.B., M.W., and J.A.K. prepared the 130nm and 28 nm CMOS samples for measurement. A. Michelson prepared the integrated circuit sample measured at NSLS-II HXN. J.J. and T.Z. implemented fly-scan controls at the APS/CNM Hard X-ray Nanoprobe. T.Z. and A.L. integrated PtychoFM workflow components at the beamline. N.S., A. Mehta, J.N.W., H.B., A.H., K.K., H.Y., M.J.C., A.R.S., S.V., and L.C.C. secured personnel, computing, and beamtime resources for this work. All authors contributed to the manuscript writing.

## Competing interests

The authors declare no competing interests.

**Figure 1: Schematic illustration of the PtychoFM model architecture and workflow.**

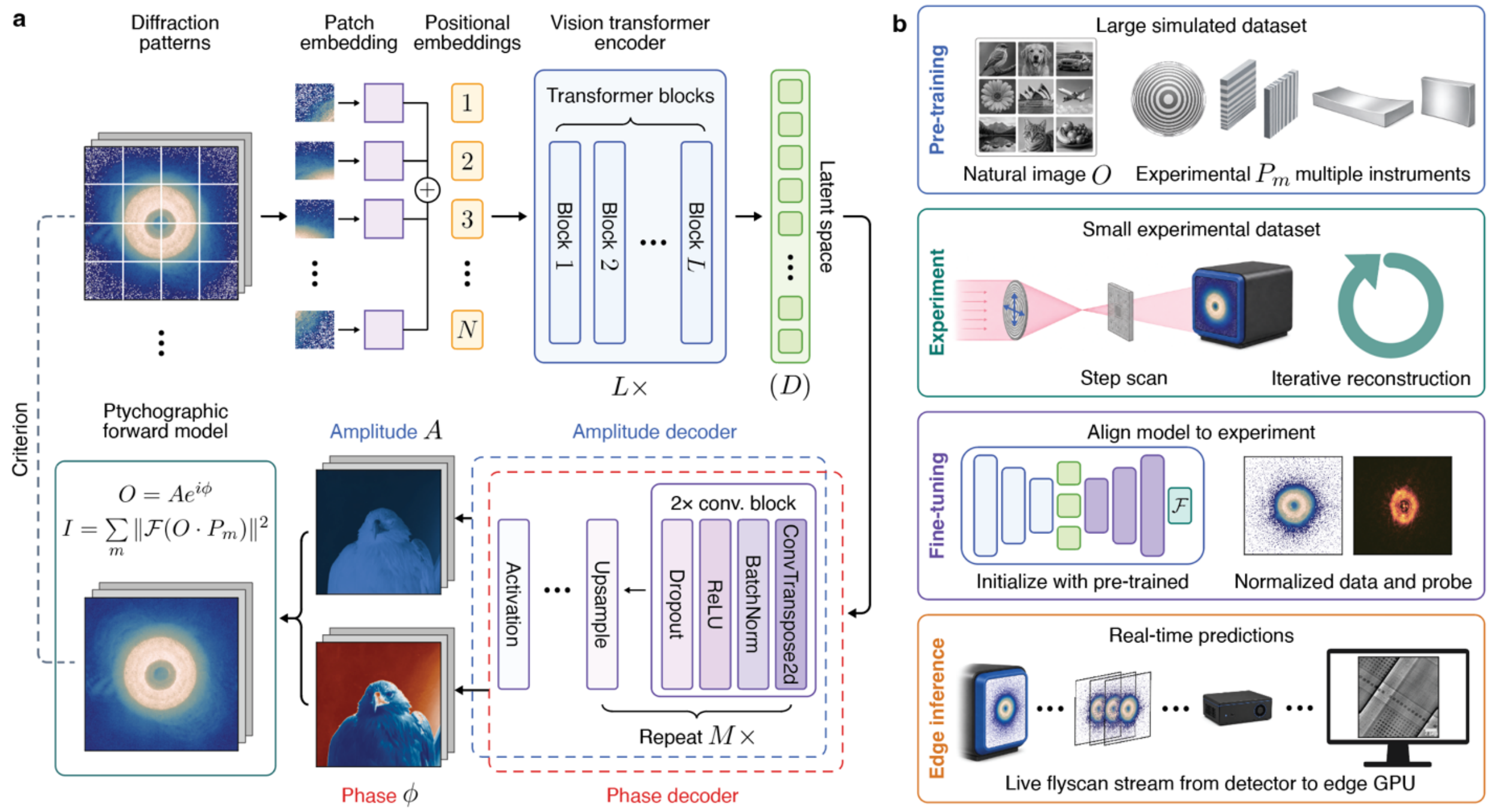

**Figure 1. a** Diagram of the physics-aware PtychoFM model architecture consisting of a vision transformer encoder and two convolutional neural network decoders. **b** Depiction of the pretraining, experimental data acquisition, fine-tuning, and edge inference workflow.

**Figure 2: Comparison of PtychoFM reconstruction of object phase to iterative methods on a 28 nm CMOS technology sample.**

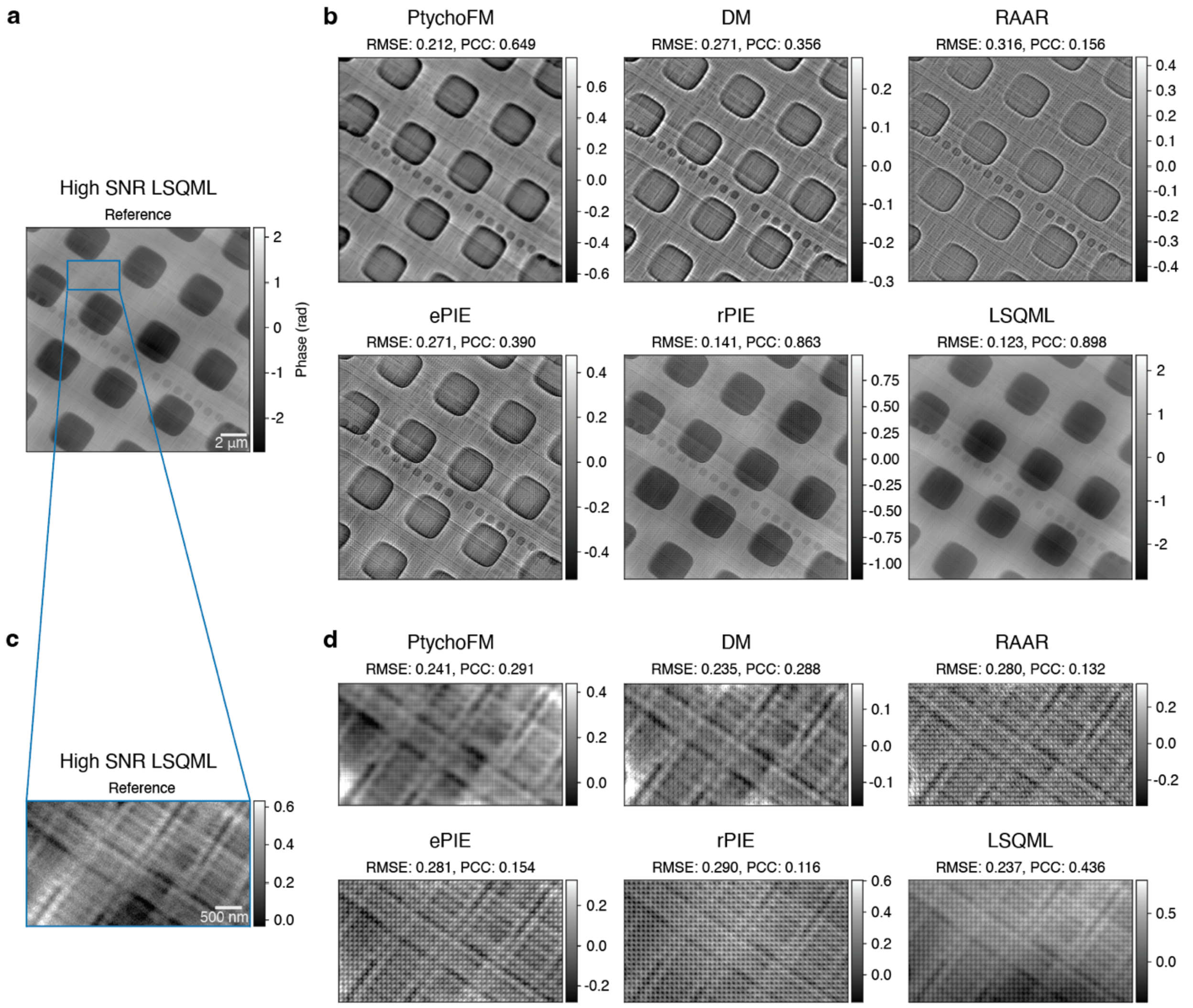

**Figure 2. a** Phase reconstruction of a high signal-to-noise ratio (SNR) step scan dataset by the least-squares maximum likelihood algorithm (LSQML). **b** PtychoFM inference, difference map (DM), relaxed averaged alternating reflections (RAAR), extended ptychographic iterative engine (ePIE), regularized PIE (rPIE), and LSQML phase reconstructions on low SNR fly-scan data collected at the same location on the sample as the reference. RMSE is the contrast-normalized root-mean-squared error and PCC is the Pearson correlation coefficient, both calculated with respect to the reference. **c** Region of interest with finer spatial features. **d** Comparison of reconstruction methods in the region of interest.

**Figure 3: Demonstration of PtychoFM workflow generalization to other ptychography instruments through comparison to iterative reconstruction.**

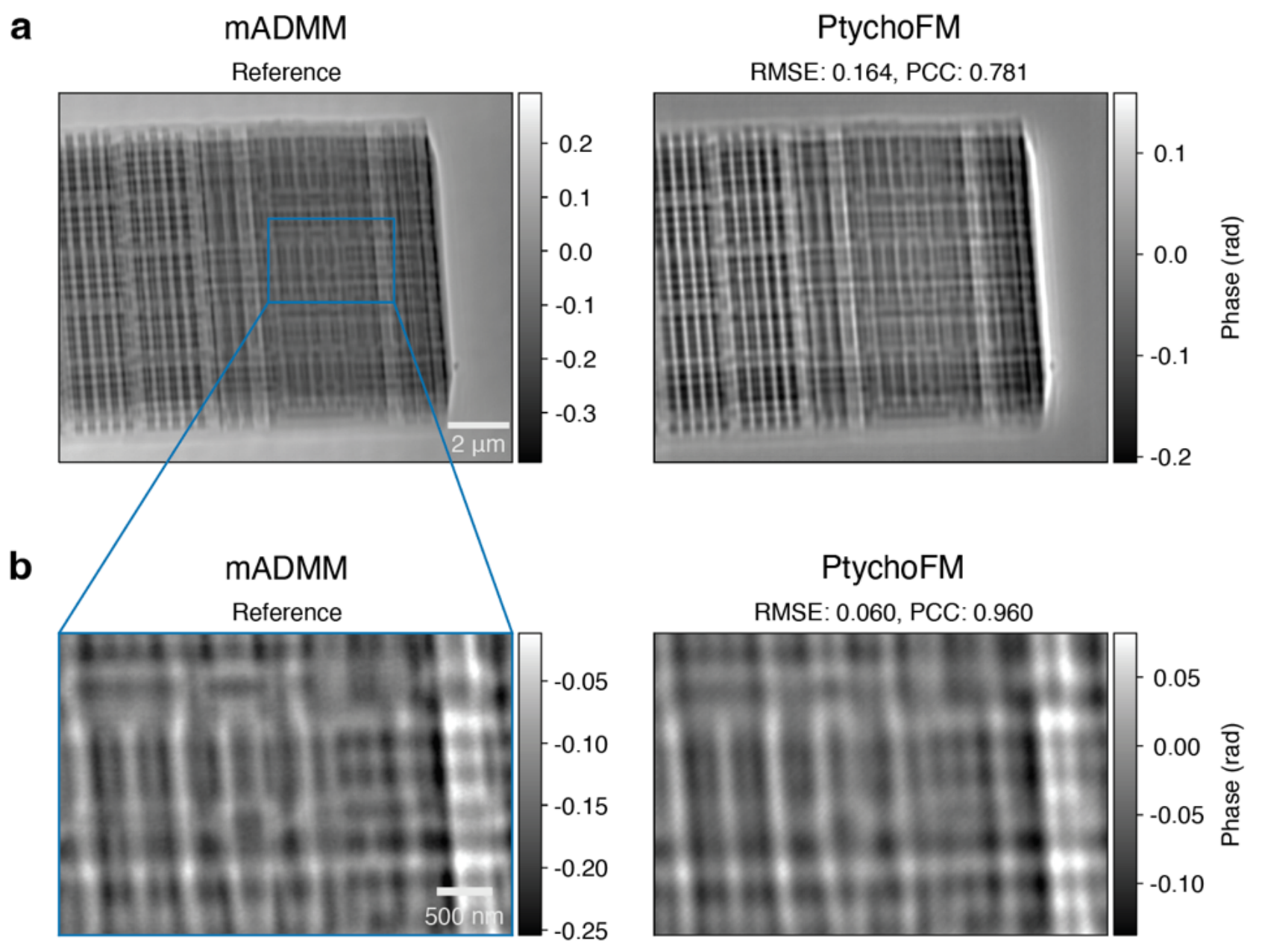

**Figure 3. a** Phase reconstruction of fly-scan data by multi-stage alternating direction method of multipliers (mADMM) and PtychoFM inference. **b** Region of interest with finer spatial features.

**Figure 4: Comparison of PtychoFM reconstruction of a Josephson junction to standard and bespoke iterative workflows.**

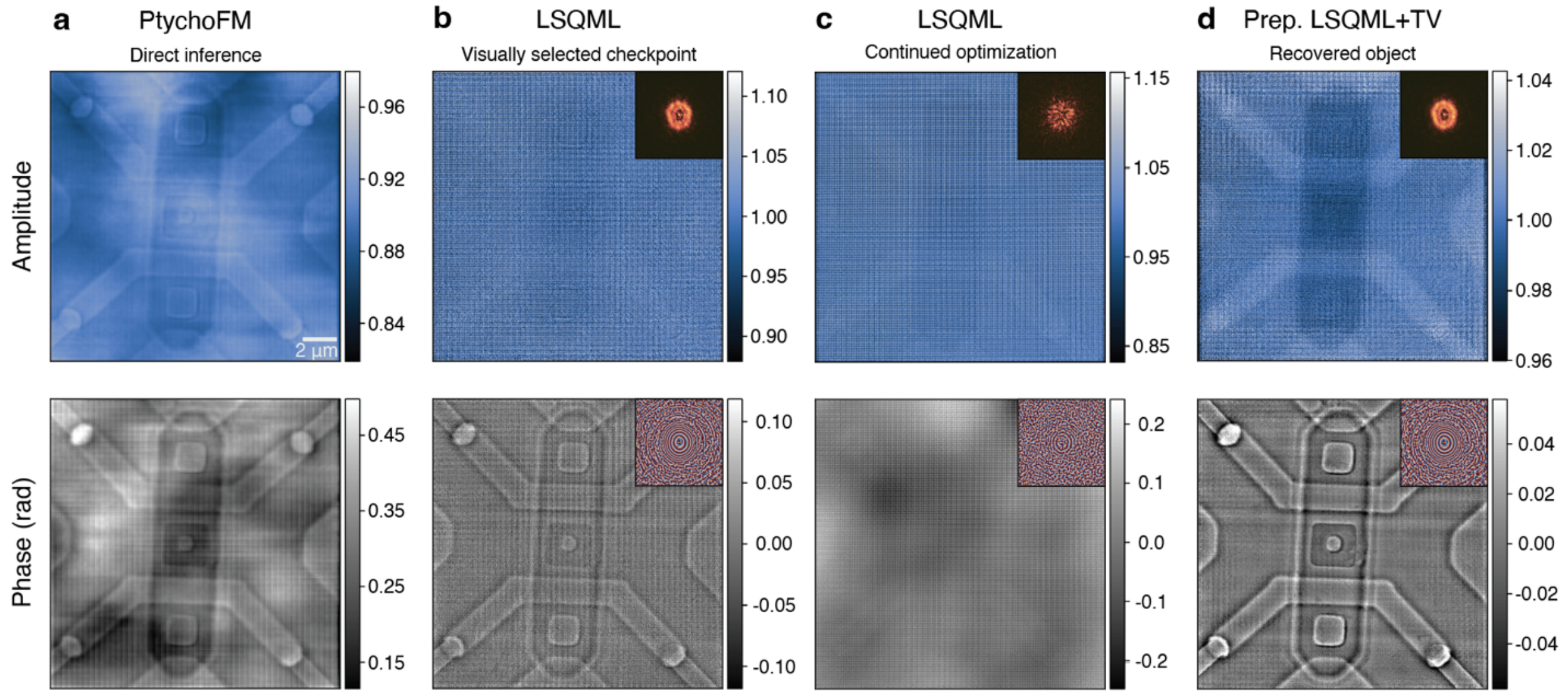

**Figure 4. a** PtychoFM prediction of object amplitude (top) and phase (bottom) from measured diffraction. **b** Visually selected reconstruction checkpoint using the least-squares maximum likelihood (LSQML) algorithm with compact batching and momentum on measured diffraction. **c** Continued optimization from (b), showing feature loss despite improved reconstruction metrics. **d** Object amplitude and phase reconstruction using a position-dependent non-uniform illumination data preprocessing step before applying the LSQML algorithm with total variation (TV) regularization. The insets in the top right corners of (b)-(d) show the amplitude and phase of the first probe mode corresponding to each respective object.

**Figure 5: Compute-optimal scaling experiments at fixed resource budgets.**

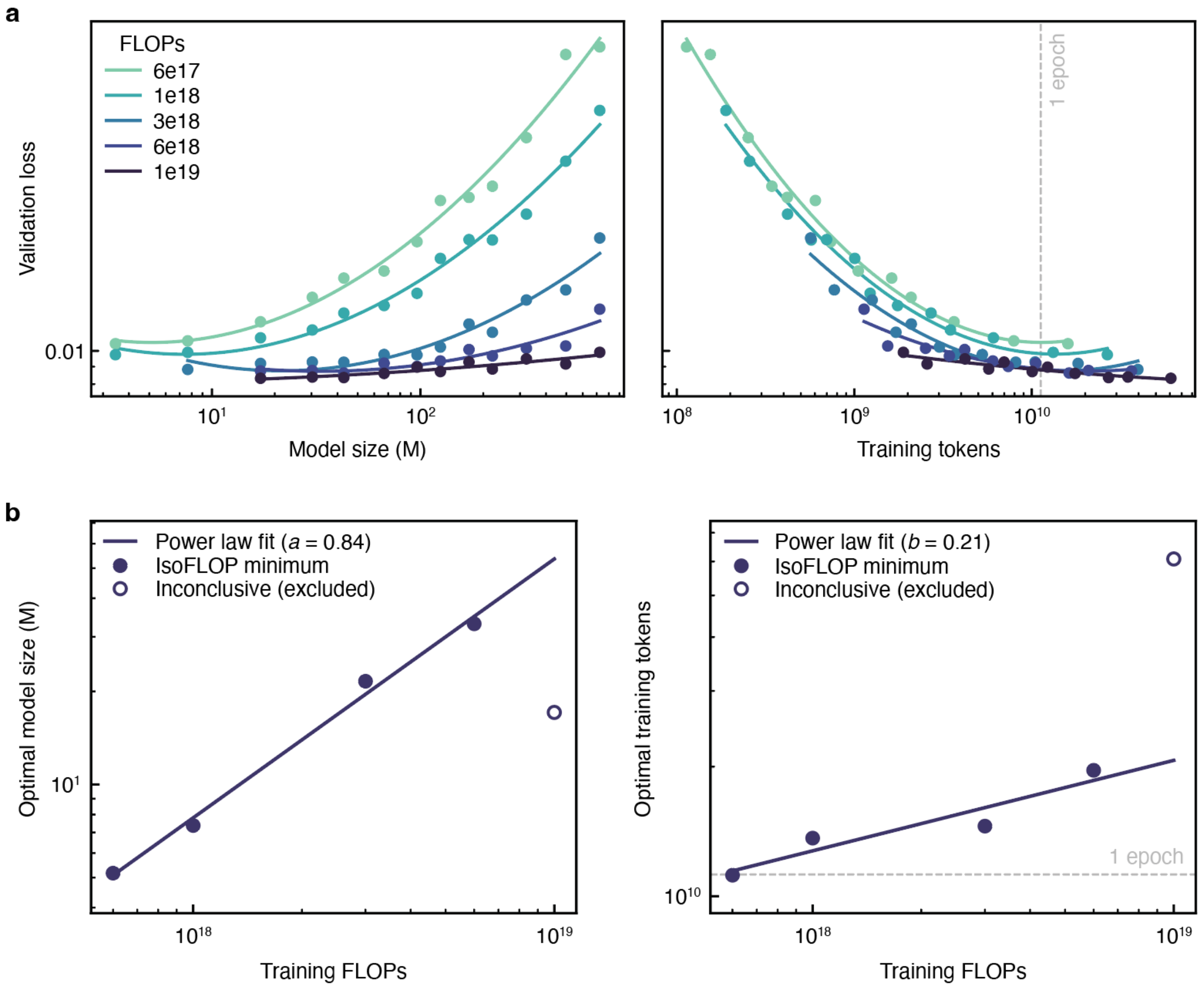

**Figure 5. a** Left: IsoFLOP log-validation loss curves fit from training different sized models described in millions of parameters (M) at fixed target compute budgets. Right: IsoFLOP curves fit from the corresponding number of data tokens used in training the models from the left panel. **b** Power law relationship fit from the minima of the isoFLOP curves in (a).

**Figure 6: Large field-of-view PtychoFM phase prediction on a 28 nm CMOS technology integrated circuit sample.**

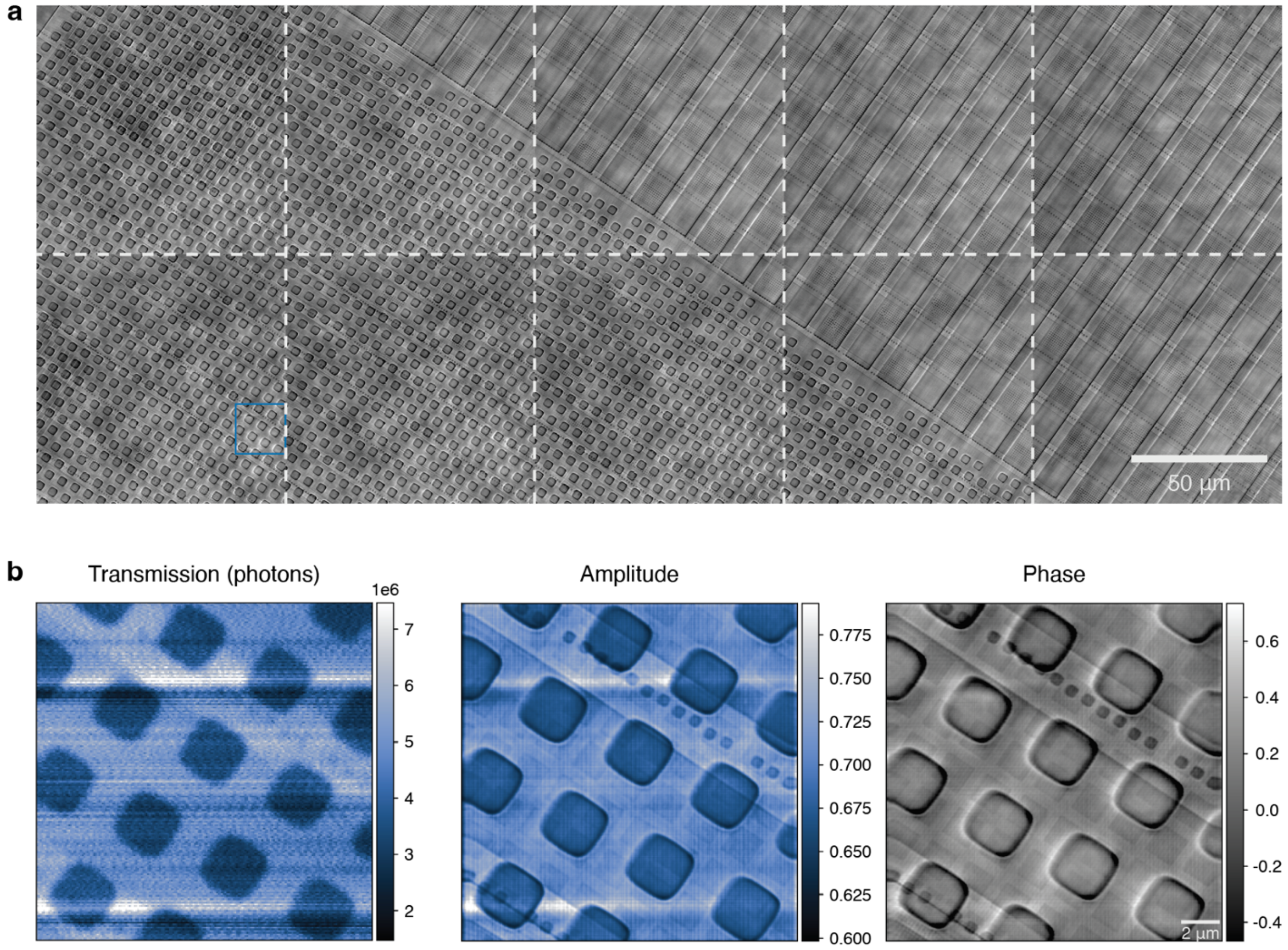

**Figure 6. a** PtychoFM phase reconstruction over an approximately 0.5 mm × 0.2 mm area (10,050,000 diffraction patterns) measured over 27 hours. The reconstruction pixel size is 6.685 nm, while the full field-of-view here is plotted downsampled by a factor of five. Even downsampled, full 300 dpi resolution would require a page width of over 45 inches to display. **b** Left: transmission contrast for selected field-of-view indicated by the blue box in (a). Middle: PtychoFM amplitude prediction showing increased resolution compared to the transmission. Right: PtychoFM phase prediction showing robustness against varying probe intensities.